\documentclass[letterpaper]{article} 
\usepackage{aaai24}  
\usepackage{times}  
\usepackage{helvet}  
\usepackage{courier}  
\usepackage[hyphens]{url}  
\usepackage{graphicx} 
\usepackage{natbib}  
\usepackage{caption} 
\usepackage{algorithm}
\usepackage{algorithmic}

\usepackage{algorithm}
\usepackage{algorithmic}
\usepackage{todonotes}
\usepackage{tabularx}
\usepackage{booktabs}
\usepackage{microtype}
\usepackage{array}

\usepackage{newfloat}
\usepackage{listings}
\DeclareCaptionStyle{ruled}{labelfont=normalfont,labelsep=colon,strut=off} 
\floatstyle{ruled}
\newfloat{listing}{tb}{lst}{}
\floatname{listing}{Listing}
\title{Coordinated Flaw Disclosure for AI: Beyond Security Vulnerabilities}

\author{
    Sven Cattell\equalcontrib\textsuperscript{\rm 1},
    Avijit Ghosh\equalcontrib\textsuperscript{\rm 2},
    Lucie-Aimée Kaffee\textsuperscript{\rm 2}
}
\affiliations {
    \textsuperscript{\rm 1}AI Village,
    \textsuperscript{\rm 2}Hugging Face,\\
    sven@nbhd.ai, avijit@huggingface.co, lucie.kaffee@huggingface.co\\
}

\usepackage{bibentry}

\begin{document}

\maketitle

\begin{abstract}
Harm reporting in Artificial Intelligence (AI) currently lacks a structured process for disclosing and addressing algorithmic flaws, relying largely on an ad-hoc approach. This contrasts sharply with the well-established Coordinated Vulnerability Disclosure (CVD) ecosystem in software security. While global efforts to establish frameworks for AI transparency and collaboration are underway, the unique challenges presented by machine learning (ML) models demand a specialized approach. To address this gap, we propose implementing a Coordinated Flaw Disclosure (CFD) framework tailored to the complexities of ML and AI issues. This paper reviews the evolution of ML disclosure practices, from ad hoc reporting to emerging participatory auditing methods, and compares them with cybersecurity norms. Our framework introduces innovations such as extended model cards, dynamic scope expansion, an independent adjudication panel, and an automated verification process. We also outline a forthcoming real-world pilot of CFD. We argue that CFD could significantly enhance public trust in AI systems. By balancing organizational and community interests, CFD aims to improve AI accountability in a rapidly evolving technological landscape.

\end{abstract}

\section{Introduction}
\label{sec:intro}

The landscape of AI accountability is rapidly evolving, as exemplified by The New York Times' lawsuit against OpenAI and Microsoft on December 27, 2023. This legal action alleges that ChatGPT was trained on the Times' published works without authorization, providing evidence of verbatim article generation \cite{nytimesTimesSues}. As part of a growing trend of litigation against AI providers \cite{apnewsPhotoGiant, authorsguildAuthorsGuild, jdsupraArtistsCopyright, acluACLUClearview}, this case highlights the complexities of replicating and verifying such claims. The Times' lawsuit has garnered significant attention, partly due to its financial resources and media reach - advantages often unavailable to smaller affected parties. This underscores how such factors can shape public discourse and potentially drive improvements in AI systems through external pressure \cite{raji2022actionable}.

The emergence of unexpected model behaviors beyond intended scope and purpose emphasizes the need for robust frameworks to identify and address these issues effectively. This stands in contrast to the well-established Common Vulnerabilities and Exposures (CVE) ecosystem in cybersecurity, where software vulnerabilities are systematically cataloged and addressed with greater clarity and consensus.

Historically, AI accountability initiatives have relied heavily on periodic audits, emphasizing repetitive assessments but lacking a structured framework for reporting user-identified issues post-deployment. Recent policy efforts have begun to acknowledge this gap \cite{altai}. This audit-centric paradigm is reflected in influential policies such as the U.S. Executive Order on AI \cite{biden2023executive}, the EU AI Act \cite{EUproposal2021}, and New York City's Local Law 144 \cite{NYC-AEDT}. However, this approach falls short when compared to the more comprehensive Coordinated Vulnerability Disclosure (CVD) processes standard in software security.

CVD plays a crucial role in enabling independent researchers to report newly identified vulnerabilities to affected vendors and the public \cite{WinNT}. This process facilitates transparent remediation before potential exploitation by malicious actors and has become a vital practice enshrined in government regulations and industry standards. Notably, the FDA mandates the implementation of CVD programs for medical device companies to enhance cybersecurity \cite{FDACybersecGuidance2023}.

While CVD has proven effective in traditional software security, its direct application to machine learning (ML) systems faces unique challenges. ML diverges from traditional software in two key aspects: 1) ML model issues must satisfy statistical validity thresholds, and 2) concerns related to trustworthiness and bias often extend beyond the typical scope of security vulnerabilities \cite{ostp_bill_of_rights}. Many critical bias reports discussed in social media and news may not align with the accepted definition of a vulnerability\footnote{The Coordinated Vulnerabilities Enumeration program, managed by MITRE and central to the CVD ecosystem, defines a vulnerability as: \textit{A problem in software, firmware, hardware, or service component resulting from a weakness that can be exploited, causing a negative impact on the confidentiality, integrity, or availability of the affected component or components.}}.

Recognizing this limitation in the conventional definition of a \textit{vulnerability}, we propose expanding the terminology with a new term - \textit{Flaw}. We define a flaw as \textit{any unexpected model behavior that is outside of the defined intent and scope of the model design}.

To address the unique challenges posed by ML systems, we advocate for adapting the CVD process into a dedicated ``Coordinated Flaw Disclosure" (CFD) framework. Tailored to ML's distinctive properties, CFD aims to formalize the recognition of valid issues in ML models through an adjudication process. This process seeks to balance the interests of vendors and the broader community, enhancing public trust by curating disclosed flaws into actionable reports. The CFD framework introduces several key innovations:

\begin{itemize}
\item Extended model cards that include detailed documentation of intent and scope, providing a baseline for flaw assessment.
\item A dynamic scope expansion mechanism that allows for the inclusion of common, unforeseen uses of the model.
\item An independent adjudication panel to mediate disputes and handle edge cases.
\item An automated verification process to streamline issue reporting and reproduction.
\end{itemize}

In the subsequent sections of this paper, we explore specific aspects of this proposal. Section 2 provides essential background information, comparing periodic audits and ad hoc harm reporting while exploring emerging participatory auditing practices. Section 3 examines evolving trends in disclosure practices within ML systems and identifies their shortcomings. Section 4 outlines the proposed changes within the CFD framework to improve disclosure rigor. Section 5 explores the practical implementation of CFD, proposing the utilization of model cards to define scope and intent. We conclude by discussing limitations of our framework and future work which includes a real-world implementation pilot and the development of a Common Use Enumeration (CUE) system to standardize the tracking of AI model applications.

\section{Background}
\label{sec:background}
In the following, we describe the background of our work. We will use the terminology of the glossary in Table~\ref{tab:glossary}.


\begin{table*}
\centering
\begin{tabularx}{\textwidth}{>{\raggedright\arraybackslash}m{3.5cm}|>{\raggedright\arraybackslash}m{\dimexpr\linewidth-3.5cm-4\tabcolsep}}
\toprule
\textbf{Concept} & \textbf{Description} \\\hline
\midrule
Vulnerability & A problem in software, firmware, hardware, or service component resulting from a weakness that can be exploited, causing a negative impact on the confidentiality, integrity, or availability of the affected component or components.\\\hline
Flaw & Any unexpected model behavior that is outside of the defined intent and scope of the model design.\\\hline
Intent & The stated purpose, goals, and intended use cases of an AI model or system, as defined by its creators or deployers. This includes the model's primary functions, target audience, and expected performance metrics.\\\hline
Scope & The boundaries of an AI model's intended operation, including what the model is designed to do, what it is not designed to do, and known limitations or exclusions. It defines the range of inputs, outputs, and use cases that are considered valid for the model.\\\hline
Coordinated Vulnerability Disclosure (CVD) & Process in which vulnerabilities will only be disclosed to the public after reporting to the software vendor and giving them sufficient time to address the vulnerability.\\\hline
Common Vulnerabilities and Exposures (CVE) & A system managed by the MITRE Corporation under the National Cybersecurity FFRDC, offering a standard method for cataloging publicly known information security vulnerabilities and exposures.\\\hline
Coordinated Flaws Disclosure (CFD) & The framework proposed in this paper to address risks derived from flaws in ML models, analogous to CVD.\\\hline
Common Flaws and Exposures (CFE) & Analogous to CVE, the proposed system to report flaws of ML models.\\\hline
Model Cards & First proposed by~\citet{Mitchell_2019}, model cards standardize the documentation of ML models, including the intended use cases of a reported model.\\
\bottomrule
\end{tabularx}
\caption{Glossary.}
\label{tab:glossary}
\end{table*}

\subsection{Coordinated Disclosure Conventions in Cybersecurity}

The cybersecurity community has cultivated a robust, 30-year history of well-established vulnerability disclosure practices, prominently featuring Coordinated Vulnerability Disclosure (CVD). This practice, acknowledged as a cornerstone of cybersecurity, facilitates collaboration between software vendors and vulnerability reporters to identify and address security flaws. Given the impossibility of eliminating all vulnerabilities during production, CVD establishes a crucial post-publication process for addressing these issues. Organizations, recognizing the limitations of in-house efforts, benefit from external parties and security researchers who disclose vulnerabilities through the CVD process. This collaboration allows affected organizations and the public to proactively address vulnerabilities, preempting exploitation by malicious actors. Moreover, the CVD process acts as a shield for those reporting vulnerabilities, demonstrating their good faith and avoiding legal conflicts \cite{CybersecCoalition2019}.

CVD's significance extends to government policies, guidance, and regulations. United States civilian government agencies \cite{CISABOD2001} and numerous federal IT contractors \cite{IoTAct2020} are mandated to adopt vulnerability disclosure policies. The Food and Drug Administration incorporates CVD into its pre- and post-market guidance for medical device security \cite{FDACybersecGuidance2023}. Key standards and guidelines from the National Institute of Standards and Technology (NIST), such as the NIST Cybersecurity Framework \cite{NISTFramework}, endorse CVD as a core practice. The EU's NIS 2 Directive \cite{EUDirective2022} requires critical infrastructure entities to establish CVD processes, and the recently passed EU Cyber Resilience Act \cite{EUCyberResilienceAct} mandates software manufacturers to adopt CVD processes. The European Union Agency for Cybersecurity further promotes national CVD legislation for its member states~\cite{CVDEUNationalPolicies}. Additionally, CVD features prominently in international standards and industry best practices on security.

CVD operates within a broader framework of programs that identify, index, and communicate security vulnerabilities, including the National Vulnerability Database, CVE system, and Common Vulnerability Scoring System (CVSS). These programs support diverse security operations such as vulnerability management, risk assessments, penetration testing, patch management, and threat intelligence \cite{CybersecCoalition2019}. CVE Numbering Authorities (CNA) are organizations external to MITRE that are authorized to issue CVEs within their scope, but MITRE acts as a ``root" that handles disputes. For example vulnerabilities in open source projects managed by the Apache Software Foundation will be sent to them first and then, if needed, appealed to MITRE \cite{apache_cna}. In the EU, the European Cybersecurity Scheme on Common Criteria (EUCC) is currently developed to cover information and communications technology (ICT) based on the international Common Criteria scheme~\cite{ENISACertification2024}.

In the realm of machine learning systems, MITRE recognizes two ML CVEs: CVE-2019-20634 \cite{CVE-2019-20634} and CVE-2023-29374 \cite{CVE-2023-29374}. While CVE-2023-29374 involves a remote code execution with an LLM exploit, the vulnerability detailed by CVE-2019-20634 in Proofpoint's system is less straightforward from a security perspective, since creating an exploitable copycat ML model requires a ton of positive and negative examples and their classification scores. Skylight Cyber's discovery of a vulnerability in Cylance's malware detection system, although awarded a Vulnerability Note by CERT, did not receive a CVE \cite{VU489481}. Evaluating the impact of these ML vulnerabilities is complex, and the existing coordinated disclosure infrastructure lacks precedence for such cases. Moreover, socioeconomic flaws in ML models \cite{solaiman2023evaluating}, not categorized as vulnerabilities (see Table~\ref{tab:glossary}), also demand reporting. 

Recognizing the importance of CVD to security, there is a temptation to extend these structures to encompass the disclosure of ML issues. However, existing CVD processes, without substantial modifications, prove insufficient to address the diverse range of ML issues. The disparities between ML models and traditional software necessitate a tailored approach. Nonetheless, the fundamental value of CVD in cybersecurity underscores the urgency of establishing a CVD-like process for ML models.

\subsection{Differences between Traditional Software and ML in Disclosure}

Traditional software vulnerabilities, such as buffer overflows, SQL injections, cross-site scripting, etc., typically arise from errors or weaknesses in the code, allowing attackers to compromise key properties like confidentiality, integrity, and availability.

In contrast to traditional software vulnerabilities with clear remediation steps, addressing ML model issues may involve retraining the model, modifying its architecture, or redefining its purpose. 
Furthermore, due to the probabilistic and complex nature of ML models, judgments are often difficult as the models lack distinct exploited and non-exploited states \cite{barbierato2024challenges}.

Most consumer ML models are deployed as closed-source APIs \cite{lunasec}, posing challenges in tracking and indexing vulnerabilities, especially in Software as a Service (SaaS) products. Versioning is not apparent to end users, and after remediation, the problem is not reproducible. Although these issues are recognized \cite{cve_for_saas}, CVEs for SaaS deployments are rare but have been issued \cite{CVE-2023-1304}. Continuous retraining and development add complexity to many AI systems.

These fundamental differences render existing vulnerability disclosure frameworks less applicable to ML model issues. Checking if an issue violates model integrity or intended functionality is less straightforward than identifying and replicating a software vulnerability due to ML models' lack of causal explanations about achieved predictions \cite{barbierato2024challenges}. In the following subsections, we delve deeper into the unique challenges posed by ML systems that existing vulnerability management frameworks cannot fully address.

\subsubsection{ML Models are Expected to Make Mistakes}

At its essence, the distinction in reporting known issues between traditional and ML-based systems lies in the expectation that ML systems will make mistakes. While these isolated incidents may be inconvenient for users, they are anticipated, and both the vendor and users should have processes in place to address this inherent limitation of machine learning. It is crucial to consider mistakes collectively, incorporating them fairly into an unbiased dataset from a statistical standpoint before deeming them as issues requiring attention.

Users anticipate that the spam model safeguarding their email will make occasional errors. A report stating that it failed to accurately classify an email provides limited utility to users and vendors, often necessitating only a minor update to the training dataset. Even when scaled up with thousands of missed emails, it may not significantly impact processes or code. The report becomes valuable when it outlines an easily executable process to circumvent the spam filter. Detailed evasion process reports may introduce operational requirements that extend beyond dataset updates and scheduled retraining, impacting the model's functionality.

However, there are instances that are entirely unacceptable to the model developer, elevating them instantly from incidents to issues. This spectrum ranges from rare, not-even-once incidents that qualify as issues to billions of data points collectively pushing the needle into issue territory. Identifying where on this spectrum an incident falls necessitates a case-by-case evaluation and is subject to various potential failures.

\subsubsection{Intent of ML Model is Hard to Quantify and Test}

Navigating ethical questions in disclosure adds complexity. Consider a model claiming demographic unbiasedness, verified by an independent auditor. If later discovered to be biased against an unexamined group, disclosure becomes crucial. Members of the affected group and users must adapt their use with this knowledge. The vendor deserves an opportunity to address and rectify the issue. Simultaneously, the independent auditor should enhance coverage for future models.

Yet, addressing ethical challenges, like bias, poses a drawback. Reports on such matters are challenging to assess. A report for a purportedly unbiased model may itself be biased, stemming from model mistakes or other biases. Distinguishing a biased report from legitimate ones becomes a challenge for external observers. While sampling biases can be refuted with a comprehensive model performance view, determining legitimacy may be impossible. Users need not alter behavior if the report is illegitimate, provided they already account for known error rates. The vendor and auditor need minimal behavioral changes, primarily preparing statements if the submitter opts to publish the report.

\subsubsection{Ethical and Safety Questions are Hard to Judge}

A pertinent example illustrating the mismatch with traditional vulnerability disclosure involves Facebook's site integrity models \cite{noorshams2020ties}. These models address issues like hate speech, spam, and harassment, impacting the system. The intricate site integrity rulebook governs the trust and safety system, encompassing machine learning models and human reviewers. The complex objective function covers socioeconomic factors for billions across diverse social groups.

Evaluating if a model's behavior violates this intricate code, distinguishing it from an understandable mistake in the moderation system, presents challenges. Submitters may create reports that, while seeming damning, fail to reveal a flaw in the moderation system. Vendors might reject valid reports on unverifiable statistical grounds, asserting the system operates within parameters. Both scenarios hinder the public – one erodes trust in an effective system, the other avoids fixing real issues.

\subsection{Absence of Unified Product or Weakness Enumeration}

Since the inception of the Common Vulnerabilities and Exposures (CVE) framework, an entire ecosystem has developed around it. In the National Vulnerability Database, a comprehensive CVE entry is now expected to encompass three key components:

\begin{enumerate}
    \item \textbf{Common Platform Enumeration (CPE)} -- Serving as a standardized naming scheme, CPE facilitates the identification of products that may contain vulnerabilities \cite{cpeoriginaldocument}.
    \item \textbf{Common Weakness Enumeration (CWE)} -- A collaborative initiative that meticulously enumerates weaknesses found in various products \cite{cwewebpage}.
    \item \textbf{Common Vulnerability Scoring System (CVSS)} -- An objective scoring system that quantifies the severity of a CVE through a numerical score \cite{cvssFirst}.
\end{enumerate}

These components, much like the CVE itself, significantly streamline the process of communicating and managing vulnerabilities within the cybersecurity landscape. The CPE allows organizations to quickly determine if their systems are susceptible to a particular CVE. Meanwhile, the CWE acts as a universal language for understanding the root causes of CVEs, thereby enhancing remediation efforts. Formally established in the mid-2000s, years after the CVE's introduction in 1999, these components have become integral to the overall CVE ecosystem.

In the realm of machine learning ethics, several independent efforts have emerged to develop a classification system for potential harms caused by ML models \cite{weidinger2022taxonomy,shelby2023sociotechnical,newman2023taxonomy,graziani2023global}. Concurrently, within the domain of ML security, initiatives like the OWASP Top 10 for Large Language Models (LLMs) \cite{owaspLLM} and ATLAS \cite{mitreMITREATLASx2122} have gained prominence. Despite these strides, the ML community has not yet reached a consensus on a universally accepted structured harm reporting taxonomy. This challenge, however, reflects a familiar trajectory seen in the evolution of the CWE project. Drawing inspiration from contributions such as the OWASP Top 10 project \cite{owasp10} and notable endeavors like PLOVER \cite{plover}, the CWE project successfully integrated diverse initiatives within focused, multi-stakeholder communities. This historical precedent suggests that the establishment of a standardized taxonomy in the ML community is indeed achievable.

\section{Disclosure Practices in Machine Learning Systems: Evolving Trends}

This section examines trends in disclosure practices for machine learning (ML) systems and the work needed to harmonize these conventions. We analyze two axes: objective security and privacy flaw disclosures, and subjective ethical flaw disclosures, such as societal bias.

\subsection{Security Disclosures for ML Systems}\label{sec:security-disclosures}

ML systems face three main types of disclosures: security vulnerability reports, model design failures, and error reports.

\subsubsection{Infrastructure Reports}

These reports highlight security vulnerabilities in ML models, often in components around the model. Examples include CVE reports on Numpy \cite{CVE-2019-6446}, Tensorflow \cite{CVE-2022-23560}, and Pytorch \cite{CVE-2023-43654}. The OpenAI bug bounty on Bugcrowd addresses infrastructure flaws \cite{bugcrowdBountyOpenAI}. SafeTensors by Hugging Face \cite{Huggingfacesafetensors} is an emerging effort to address reproducibility concerns.

An instance of a traditional vulnerability is DeepSloth \cite{deepsloth2020}, a Denial Of Service attack on LLMs. ProofPudding \cite{nistCVE201920634} was the first CVE issued for an issue caused by an ML model. The mitigation involved limiting API information \cite{oksuz2023autolycus}.

\subsubsection{Model Design Failure Reports}

Model design failures manipulate samples to induce changes in output. Unlike infrastructure reports, they rely on modifying model output. Examples include the ``Bayesian Poisoning Attack" \cite{bayesian_poisoining} and the Cylance malware detector \cite{cylanceikillyou}. The Malware Detection Evasion Competition \cite{malware_evasion_competition} highlights such vulnerabilities.

Adversarial examples are often considered low priority \cite{szegedy2014intriguing}. They can be defeated by simple techniques like jpeg compression \cite{aydemir2018effects} or blur \& repair \cite{carlini2022certified}. While adversarial patches \cite{brown2018adversarial} are possible, they are hard to execute \cite{athalye2018synthesizing}. Marines bypassed an ML model with a simple attack \cite{marines_box}, highlighting that lower-hanging fruit is often more appealing \cite{apruzzese2022real}.

\subsubsection{Error Reports}

Error reports stem from expected model errors or systematic errors amendable through data changes. These include:

\begin{itemize}
    \item \textbf{One-off Errors:} Normal errors within the model’s expected error rate.
    \item \textbf{Systematic Errors:} Errors from dataset biases, leading to harmful outputs. Examples include the Google image classifier’s mislabeling \cite{google_gorrilla} and the ChatGPT divergence attack \cite{nasr2023scalable}.
\end{itemize}

\paragraph{Reports that Depend on Context}

Recent reports include prompt injection attacks \cite{nvidiaSecuringSystems}, with vulnerabilities in LangChain receiving CVEs. These can be “model design failure reports” or “error reports” depending on the context. For instance, basic chatbots might not be vulnerable, but complex chatbots must have measures to address these risks \cite{Burgess_2023}. Proper disclosure is crucial for systems granting sensitive access, such as email or calendar functions.

\subsection{Disclosure of Ethical Issues in ML Systems} \label{sec:ethics}

Despite calls for audits from various stakeholders \cite{raji2022outsider,fjeld2020principled,wilkinson2023audits}, the identification and disclosure of ethical issues in machine learning (ML) systems have been ad hoc and reactive. Unlike traditional software systems with established practices for coordinated vulnerability disclosure, ML model issues often surface inconsistently. Without a mandated reporting mechanism, harms caused by flawed ML systems are often discovered post-damage \cite{google_gorrilla}. Few proactive mechanisms exist for auditing these complex systems, identifying pre-deployment risks, and enabling responsible disclosure balancing organizational and community interests. However, approaches like cooperative audits \cite{wilson2021building}, participatory audits \cite{dennler2023bound}, and AI red teaming \cite{OpenAI2023,ganguli2022red} are emerging as ML stakes grow. This section traces the history and trends in disclosing ethical issues in ML.

Reporting has primarily relied on affected communities raising complaints through social media, media coverage, lawsuits, and public pressure after issues arise. Few systematic mechanisms audit and identify model issues proactively. Examples include the racial bias in COMPAS recidivism prediction algorithms uncovered by investigative journalism \cite{ProPublicaMachineBias} and facial analysis systems misclassifying people of color until public backlash erupts \cite{BBCGoogleRacistBlunder}. Harmful biases in language models generating toxic outputs against marginalized groups are often observed due to vigilance within directly impacted communities \cite{CosmosMagazine2023,abid2021persistent}.

The lack of coordinated discovery and disclosure processes and the closed-source nature of many AI systems mean users remain unaware of problems until they surface. Reactive reporting by harmed parties has made accountability an exception rather than the norm for ML systems. Researchers and practitioners emphasize the need for proactive, community-centered mechanisms for responsible disclosures. Providing legal protections for independent ML issue researchers, akin to protections for good faith security research, is essential \cite{HPCAI2023}.

\subsection{Participatory Approaches for Issue Discovery}

In response to growing scrutiny around AI ethics and the limitations of ad hoc reporting, organizations are experimenting with participatory approaches to systematically discover model issues.

\subsubsection{Bias Bounty Programs}

Bias bounty programs incentivize crowd-sourced identification of algorithmic problems by offering rewards and recognition for impactful findings. Early efforts include Twitter’s algorithmic cropping bounty \cite{yee2021image} and QueerInAI’s queer bias bounty \cite{dennler2023bound}. AI bug bounties are becoming common in large companies \cite{googleblogGooglesReward,microsoftMicrosoftBounty}.

\subsubsection{Community Driven Audits}

Community-driven audits, petitions, and pressure campaigns are effective for advocacy groups to highlight urgent ML issues and hold organizations accountable \cite{AIVillageRedTeam2023,NYT-AI-Hackers2023}. Truly participatory approaches require investments in inclusion to lower barriers for directly affected groups \cite{dennler2023bound}. Transparency in model development and deployment fosters constructive feedback. However, relying on ML model owners for authorization can undermine researchers' independence \cite{DataSocietyRedTeaming2023}.

\subsubsection{Crowdsourced AI Harm Databases}

Standardized documentation of AI harms is nascent. Efforts like MITRE's ATLAS\footnote{\url{https://atlas.mitre.org/}}, AI Vulnerability Database (AVID)\footnote{\url{https://avidml.org/}}, and AI Incident Database (AIID)\footnote{\url{https://incidentdatabase.ai/}} are well-known. AIID collects news stories about AI incidents, while AVID assesses open-source models. These initiatives lack coordinated disclosure features and vendor rebuttal mechanisms. Bounty programs like Google's and Microsoft's show an embrace of disclosure at large model vendors \cite{googleblogGooglesReward,microsoftMicrosoftBounty}.

\subsubsection{AI Red Teaming}

AI red teaming involves organizations commissioning external experts to rigorously test and attempt to attack their ML systems, simulating real-world malicious actors \cite{OpenAI2023,ganguli2022red}. Goals include pressure-testing models, surfacing overlooked issues, validating internal safety claims, and gaining new perspectives to strengthen systems before public release \cite{selbst2021institutional,DataSocietyAlgorithmicImpact2021,carlini2023aligned,zou2023universal}. Effective red teams require diverse professional backgrounds, ideologies, regional representations, and skill sets.

While promising, these community-centered mechanisms are not yet the norm. Much work remains in formalizing scalable, sustainable, and equitable participatory practices for discovering and disclosing ML issues.

\section{Proposed Changes to Recognition of a Disclosure}

In traditional coordinated disclosure for software vulnerabilities, if a vendor rejects acknowledging a reported issue, the only recourse for the submitter is to publicly disclose the problem and advocate for recognition. While this can be effective in drawing attention to and resolving vulnerabilities, the dynamics change when dealing with ML-based systems.

ML disclosures often involve ethical considerations, particularly regarding political speech. The challenge lies in establishing a process that fairly evaluates submissions, distinguishing substantiated claims from unsupported ones aiming to serve the submitter's agenda. Accepting ethical issue reports based on sampling bias can erode public trust in the vendor, while rejecting valid issues supported by statistically sound data damages both public trust in the vendor and the entity responsible for the flawed rejection.

To address these challenges, we propose a Coordinated Flaw Disclosure (CFD) process that incorporates:

\begin{enumerate}
    \item \textbf{Extension of Model Cards:} Model Cards, traditionally used for complete systems, will be extended to provide detailed documentation of intent and scope.
    
    \item \textbf{Trusted Independent Adjudication Panel:} A panel of independent adjudicators will be established to mediate disputes between the submitter and vendor when seeking official recognition.
    
    \item \textbf{Automated Verification through Submission Extension:} An extension of the submission concept will enable automatic verification of the potential issue's existence, enhancing the efficiency of the recognition process.
\end{enumerate}

The main difference between this and the current CVE process is that it is based on the specific model or system card, not a universal definition of vulnerability. Bug bounty programs usually have a ``scope" document associated with each vendor that dictates what is and is not reportable. The CVE process does not have this pre-defined scopes for a system; the definition of the vulnerability and community norms manage it. Vendors may argue for a lower CVSS score from MITRE if they believe the vulnerability is not relevant. However, there are flaws that are reputable in one model but not in another due to the creator's intent. Llama2 has an extensive model card \cite{llama2_model}, but Mistral is meant to be a base model that is fine-tuned and embedded in another system, and the model card reflects that \cite{mistral_model}. Mistral's card is quite sparse and incomplete for a process such as this.

The main component is the trusted independent adjudication panel to govern the disputes that will arise between the submitter and vendor. This is a key part of the official CVD processes, as demonstrated by the ``root'' in the CVE Numbering Authority (CNA) rules for 3rd party CVE enumeration \cite{cna_cve}, the Bugcrowd appeals process \cite{bugcrowd_make_it_right}, and the HackerOne Mediation process \cite{h1_make_it_right}. This will provide an outlet that fairly hears the submitter but also the vendor. If there’s sensitive proprietary information from the vendor that refutes the issue, the adjudicator can review it and resolve the dispute without a public fight. However, if this data does not truly dispute the submitter, it can award the issue and thus start the mitigation process at the vendor.

We hope that this process will enable vendors to create safer and more secure models, and the public to understand them better. We expect the discovered issues that arise in the course of a model system’s development to impact its Model Card to make it more precise, thus enabling everyone to better understand the capabilities of the system.

\subsection{Complete Model/System Card}

The vendor is required to provide a comprehensive model card for the entire system, detailing both the intent and scope. The intent encompasses the system's purpose and objectives, while the scope delineates the limitations and known issues associated with the system. This requirement aligns with the guidance established in the seminal work by Mitchell et al. \cite{Mitchell_2019} and is exemplified in various model cards, such as the Llama2 card \cite{llama2_model}.

The intent section should include a clear mission statement for the system and outline the efficacy measures relevant to the model. These elements are expected to evolve as the model's performance and its contextual understanding improve. The scope section should address both current issues being addressed and those that are unresolved within a reasonable timeframe. These unresolved issues may be categorized into fundamental flaws that cannot be rectified and known flaws that are under ongoing development but may take several years to resolve.

This approach effectively distinguishes between two aspects of ``scope" commonly used in the security domain. While ``intent" encompasses the inherent fallibility of the system, ``scope" in security contexts often refers to vulnerabilities within the specified parameters. The intent is generally understood through product descriptions and community feedback. Given that models with identical input and output formats may serve different purposes, a clearly defined intent statement helps clarify the system's objectives and minimizes the need for extensive scope details. For instance, the Microsoft bounty program illustrates this distinction by excluding prompt injections from scope unless they access privileged information from other users.

A complete system card encompasses all components of the system, including both vulnerable models and those designed to mitigate such vulnerabilities. For example, integrating a diffusion model with an image classifier might address adversarial examples, with the model card specifying this capability as a system property rather than an attribute of the individual model \cite{florian_robustness}. In this scenario, adversarial examples would be considered ``in-scope" and reportable. Similarly, in the case of an LLM, deploying a hate speech detection model alongside the LLM could prevent the generation of hate speech, with the LLM model card noting the potential for hate speech, while the system card indicates that hate speech is effectively blocked. Another example includes Dalle2's use of demographic descriptors to enhance output diversity \cite{dalle2_bias}.

We now further elaborate on what ``Intent" and ``Scope" mean in the context of our proposed framework.

\subsubsection{Intent}

To make ``intent" actionable, the system card should include a mission statement for the model system, and as many precise efficacy measures as possible. 

Examples of Intent Mission statements:
\begin{enumerate}
    \item From a text prompt, we produce images that are safe for work, safe for children, and free of demographic bias.
    \item This malware model classifies Portable Executable (PE) binaries into malicious or benign. We consider Potentially Unwanted Applications, such as adware, to be malicious.
\end{enumerate}

Examples of Intent measures:
\begin{enumerate}
\item We reproduce images of professionals in particular careers at the demographic rates of the US in this Text to Image model.
\item We have a 0.001 False Positive Rate and a 0.1 False Negative Rate on each customer’s cumulative data for this malware model, if that customer has over 100 endpoints.
\end{enumerate}

\subsubsection{Scope}
The main purpose of Scope is to exclude known issues that the vendor has no intent or ability to resolve in the next year. There are intents, like Mistral's release of a base model for use in a larger system, that essentially exclude many reportable flaws. The two scope documents from the Google and Microsoft AI Bounty programs provide good examples of in and out-of-scope statements.

\begin{enumerate}
\item L2 norm adversarial examples are a known vulnerability in this image model; do not use for sensitive applications.
\item This is an LLM with no protections; prompt injections are out of scope.
\item Binary packing is a known vulnerability of this malware detection model, and we have no plans to build mitigations.
\end{enumerate}

\subsection{Adjudication Process}

The vendor should not have sole authority over the recognition of an issue, especially when they are the only entity that can verify the issue's existence. Additionally, reports should be carefully vetted to ensure that the issue truly exists, is a valid violation of the model card, and is not a result of sampling bias in the provided report samples. We acknowledge that this is a significant deviation from the CNA process used to grant vendor CVEs and anticipate that it will be a topic of discussion.

The primary function of this committee is to distinguish between issues that involve valid violations of the model’s intent and scope and those that do not. One challenge arises when an unbiased model used in hiring or image generation may not precisely align with a demographic split. If the model card does not allow for an error in this measurement, it would be considered a violation in the strictest sense. However, if the error is within industry standards, a report detailing this may not be officially recognized.

There are instances where the decisions made by the model may violate one of the stated intents of the model but align with another. Achieving complete political unbiasedness may be challenging, especially if one political group violates another policy far more frequently than another group.

\subsection{Automatic Verification}

Model errors are anticipated to constitute a significant proportion of reports in any database of AI system issues. A structured reporting mechanism, such as a telemetry system, offers several advantages to vendors. Primarily, it facilitates seamless integration of reported issues into the model's training and testing datasets. This approach also ensures that the reported behavior is verifiable and reproducible—a critical consideration given the potential for daily fluctuations due to model updates and continuous training processes.

The majority of substantive reports, as outlined in the background section, include demonstrative samples of the problematic behavior. Therefore, any system supporting Coordinated Flaw Disclosure should incorporate functionalities for automatic external verification and integration into the vendor's internal testing dataset. This approach significantly reduces the vendor's resource expenditure in issue remediation and provides a mechanism for ongoing verification of issue resolution.

Upon acceptance of a reported issue, the system should preserve the supporting data and implement periodic re-verification protocols. This is particularly crucial as issue resolution in one model iteration does not guarantee its absence in subsequent versions, potentially leading to regression and issue recurrence. By maintaining a robust verification system, vendors can ensure continued model integrity and swiftly address any re-emerging issues.

\begin{figure}
    \centering
    \includegraphics[width=0.7\columnwidth]{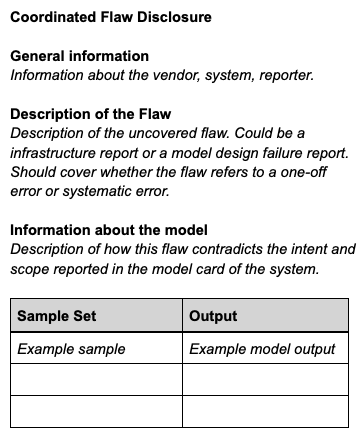}
    \caption{A potential template for a CFE report by a user.}
    \label{fig:submission-template}
\end{figure}

\subsection{Dynamic Scope Expansion for Common Use} \label{sec:commonuse}

For general-purpose models\footnote{also called foundation models or frontier models, some of these models are additionally open-weight with permissive licensing and have led to an ecosystem of downstream use cases, such as Meta's LLAMA family of models (\url{https://llama.meta.com/}).} with potentially limitless applications, we propose a ``Common Use'' clause in the model card. This clause allows for the dynamic expansion of the model's intent and scope based on widespread adoption of certain use cases, even if they were not originally defined.

Initially, individual companies should strive to identify and incorporate common uses of their models. However, as the field matures, this process could evolve into a centralized effort, potentially managed by a cross-vendor coalition or an independent body. Such an approach could lead to the development of a Common Use Enumeration (CUE), complementing existing systems like the Common Weakness Enumeration (CWE).

To implement this approach:

\begin{enumerate}
    \item \textbf{Define a threshold for ``common use'':} Establish quantitative criteria (e.g., X\% of users or Y number of deployments) to determine when a use case becomes 'common'.
    
    \item \textbf{Regular review process:} Implement a periodic review (e.g., quarterly or bi-annually) to identify emerging common uses of the model.
    
    \item \textbf{Scope expansion:} For uses meeting the defined threshold, expand the model card to include these as in-scope for flaw reporting and mitigation.
    
    \item \textbf{Allow reporting:} Enable the submission of issues related to these expanded use cases through the CFD process.
\end{enumerate}

The Common Use clause acknowledges that while general-purpose models may have vast potential applications, safety and responsible development require some boundaries. By allowing for dynamic scope expansion, we create a framework that can evolve with the actual use patterns of the model, enhancing the overall effectiveness of the CFD process. As the AI field continues to advance, the development of a centralized, independent CUE system could further streamline this process and promote industry-wide consistency in addressing common uses of AI models. We further discuss a CUE system in the Future Work section of this paper.

\section{Issue Reporting Governance}

In the following, we describe the process of submitting a CFD. The implementation of the process depends on the vendors, however, their clear advantage of formalizing the process is similar to the adoption of the previous CVD process; vulnerabilities are explored before they can be exploited. Where CFD-related issues are identified, we can assume that it is in the best interests of the vendors to be notified about those to be able to resolve them in a timely manner. Our proposed CFE issuance process (Section~\ref{sec:issuance}) takes this into account, and proposes a process  to resolve potential issues between reporters and vendors.

\subsection{Submitter Interaction}

Users identify potential issues believed to violate the scope and create a submission with:
\begin{enumerate}
    \item The sample set (even a single sample is acceptable)
    \item The output(s) the model should not generate.
\end{enumerate}

The report argues that the sample output pairs violate the scope. After submission, the report is sent to the vendor for acceptance or rejection. If rejected, the user can appeal to the Adjudicators.

We propose a reporting template in Figure~\ref{fig:submission-template}. The submitter should provide information about the vendor, system, and the person reporting the issue. The description of the uncovered flaw should include whether this is a infrastructure or model design failure report, and whether it is a one-off error, or a systematic error. For more details on the error types, see Section~\ref{sec:security-disclosures}. 
To give the vendor context about how to address the issue and a possibility to prioritize the report, the report should refer to the model card of the model. 
Finally, the template should have the possibility to submit sample set and according output, depending on the size either as a table or as structured in a common data format.

\subsection{CFE Issuance Process}\label{sec:issuance}
\begin{figure}
    \centering
    \includegraphics[width=0.7\columnwidth]{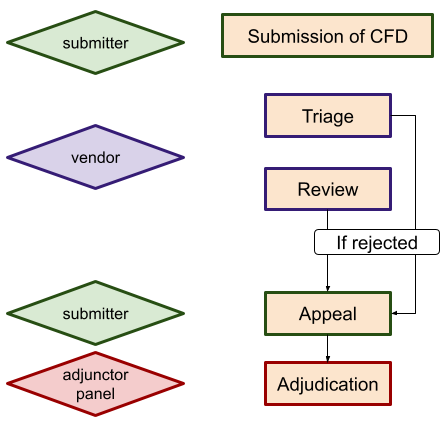}
    \caption{Flowchart of the CFE issuance processes described in Section~\ref{sec:issuance}.}
    \label{fig:flowchart-cfe}
\end{figure}

An overview of the CFE issuance process can be seen in Figure~\ref{fig:flowchart-cfe}.
Upon submission of a flaw, the following steps are followed:

\begin{enumerate}
    \item \textbf{Triage}: The vendor or an authorized person reviews the report to ensure the sample and output pairs satisfy the report.
    \begin{itemize}
        \item[] \textbf{Note:} If not, the submission is rejected on technical grounds.
    \end{itemize}
    \item \textbf{Review}: If the report and sample/output pairs align, the vendor determines if the report is within Scope and violates the model Intent.
    \begin{itemize}
        \item[] \textbf{Note:} If not, the submission is rejected as either out-of-scope or in-intent.
        \item[] \textbf{Caveat:} If the data supports the report but is incomplete or flawed, the vendor can reject on statistical grounds or request additional data.
    \end{itemize}
    \item \textbf{Adjudication}: If rejected for any reason, the Submitter can appeal to the adjudicator panel.
\end{enumerate}

The vendor may reject the submission on statistical grounds to the adjudicator panel. If the submitter-supplied sample/output pairs are statistically biased, and an unbiased sample set would not show a violation of the model card, the adjudicator panel may request supporting data from the vendor to validate the rejection. The adjudicator and vendor can jointly assess whether the vendor-supplied data is too sensitive for the submitter and decide on further steps. An example of this cooperation is once again the ``divergence attack'' paper by Google \cite{nasr2023scalable}. The authors followed a standard 90 day responsible disclosure practice\footnote{\url{https://not-just-memorization.github.io/extracting-training-data-from-chatgpt.html#responsible-disclosure}} in their private disclosure of the bug to OpenAI and publicly reported the bug only after the 90 days were up.

\subsection{Adjudication Process for Edge Cases}

While the CFD process primarily focuses on issues within the defined scope and intent of a model, the dynamic nature of AI systems, particularly general-purpose models, necessitates a mechanism for addressing edge cases. These edge cases often arise from common uses that were not initially anticipated in the model card. To address this, we propose an extended adjudication process:

\begin{enumerate}
    \item \textbf{Initial Triage:} When a report is submitted that falls outside the stated intent but appears to be within common use, it is flagged for special review.
    
    \item \textbf{Common Use Assessment:} The adjudication panel first determines if the reported use case meets the criteria for 'common use' as defined in the Dynamic Scope Expansion clause (Section \ref{sec:commonuse}).
    
    \item \textbf{Harm Evaluation:} If the use case is deemed common, the panel assesses the severity and prevalence of the reported harm.
    
    \item \textbf{Vendor Consultation:} The panel consults with the vendor to understand any technical constraints or considerations related to the reported issue.
    
    \item \textbf{Decision:} Based on these factors, the panel may:
    \begin{itemize}
        \item Recommend immediate inclusion of the issue in the CFD process
        \item Suggest updating the model card to include this use case in future versions
        \item Advise on interim mitigation strategies
        \item Determine that the issue, while outside of scope, warrants public disclosure due to its potential impact
    \end{itemize}
    
    \item \textbf{Follow-up:} The panel ensures that any decisions are implemented, including updates to the model card, CFD database, or public disclosures as necessary.
\end{enumerate}

This extended process balances the need for clear boundaries with the reality of evolving AI applications, providing a structured approach to handling edge cases that may otherwise fall through the cracks of traditional disclosure processes.

\subsection{Adjudicator Panel}
Current policy proposals focus on providing audits by external experts~\cite{DBLP:journals/corr/abs-2403-07904}. These audits have a very wide range of tasks and possible restrictions to apply. In the CFD process, these already planned panels could also act as adjudicator panels in the CFD processes. It would give the audits a starting point about contested issues, and could be a starting point for the implementation of the CFD process. It is important to emphasise that we expect the vast majority of CFD to be resolved between the reporters and AI providers directly as it is the case with the CVE processes \cite{redhat_rootcna,cvedrama}, keeping the workload of the adjudicator panel light. 

\section{Conclusion}

This paper introduces a Coordinated Flaw Disclosure (CFD) framework for machine learning and AI systems, addressing algorithmic flaws through transparent intent and scope definitions via extended Model Cards. The CFD framework balances submitter and vendor interests, fostering trust and transparency through structured issue reporting governance. By considering both quantitative and qualitative evidence, it captures a holistic view of AI system flaws. We believe widespread adoption of this framework can bolster public trust in AI systems by promoting clear processes and collaborative engagement. As machine learning evolves, the CFD framework offers a structured, adaptable approach to navigate the changing landscape of algorithmic flaws, benefiting both vendors and the public.

\subsection{Future Work}

While our proposed CFD framework lays the groundwork for improving AI accountability and transparency, there are several areas where further development and real-world testing are needed. We outline two key areas for future work: an implementation pilot to test the CFD framework in practice, and the development of a Common Use Enumeration system to standardize the tracking of AI model applications.

\subsubsection{Implementation Pilot at DEF CON 32}

In collaboration with key stakeholders from government, industry, freelance security researchers, and policy experts, we are organizing the Generative Red Team 2 (GRT2)\footnote{\url{https://grt.aivillage.org/announcement}} event at DEF CON 32. This event serves as a comprehensive test of the CFD framework's robustness and applicability in real-world scenarios. Key aspects of GRT2 that align with and test our CFD framework include:

\begin{itemize}
    \item \textbf{Model Cards:} Participants receive detailed model cards specifying the intent and scope of the target AI system, mirroring our proposed extended documentation approach.

    \item \textbf{Comprehensive Flaw Reports:} Using the UK AI Safety Institute's Inspect AI framework\footnote{\url{https://inspect.ai-safety-institute.org.uk/}}, participants prepare statistically valid reports on model flaws, moving beyond single-example vulnerabilities.

    \item \textbf{Dedicated Submission Platform:} Reports are submitted through a platform, demonstrating the feasibility of specialized CFD infrastructure.

    \item \textbf{Vendor Interaction:} AI system vendors review and respond to submitted reports, simulating real-world disclosure dynamics.

    \item \textbf{Independent Adjudication:} A panel of independent experts mediates disputes between submitters and vendors, testing the efficacy of our proposed adjudication process.

    \item \textbf{Transparency:} All data generated during the event will be made public, aligning with our goals for increased transparency in AI flaw disclosure.
\end{itemize}

GRT2 is designed not only to validate the CFD framework but also to stress-test it, identifying potential weaknesses and areas for improvement. The results and lessons learned from GRT2 will be invaluable in refining the CFD framework.

\subsubsection{Common Use Enumeration (CUE) System}

To address the challenge of tracking and standardizing common uses of AI models, we propose the development of a Common Use Enumeration (CUE) system. This system would serve as a centralized repository for documenting and categorizing the various applications of AI models, particularly focusing on general-purpose or foundation models that may have a wide range of unforeseen uses.

\paragraph{Proposed Structure of the CUE System}

\begin{enumerate}
    \item \textbf{Hierarchical Classification:} The CUE system would organize AI model uses into a hierarchical structure, similar to the Common Weakness Enumeration (CWE) system. This would allow for broad categories (e.g., ``Natural Language Processing", ``Computer Vision") with increasingly specific subcategories.

    \item \textbf{Unique Identifiers:} Each documented use case would receive a unique identifier (e.g., CUE-001) for easy reference and tracking.

    \item \textbf{Metadata Fields:} Each entry would include relevant metadata such as description of the use case, associated model types, potential risks or ethical considerations, known implementations or case studies, and related CUEs or CWEs.

    \item \textbf{Version Control:} The system would maintain version history to track how use cases evolve over time.
\end{enumerate}

\paragraph{Management and Governance}

We propose the establishment of a multi-stakeholder governing body to oversee the CUE system, modeled after established multi-stakeholder processes in the CVE ecosystem. This body would manage a standardized submission process, regular updates, and ensure open access to the CUE database. Vendors could integrate CUE identifiers into their model cards, use the system for reporting new uses and risk assessment, and access an API for integrating CUE data into their development and monitoring tools.

The development of a Common Use Enumeration system represents a significant undertaking that would require collaboration across the AI industry, academia, and regulatory bodies. However, such a system could greatly enhance our ability to track, understand, and responsibly manage the diverse applications of AI models in society.

\subsection{Limitations}

While the CFD framework presents a significant step forward, several challenges remain:

\begin{itemize}
    \item \textbf{Model Cards Standardization:} The lack of standardized model cards poses a challenge for widespread CFD adoption. Establishing standards for model cards is crucial for effective implementation, as emphasized by the NIST AI Risk Management Framework \cite{nist_model_cards}.
    
    \item \textbf{Dynamic ML Datasets:} Tracking and managing evolving datasets used in model training presents complex challenges. Unlike software dependencies tracked by SBOMs \cite{sboms,sboms_josh}, monitoring dynamic ML datasets lacks established tools. Recent incidents, such as the LAION-5B dataset retraction \cite{schuhmann2022laion5b,laion_csam,csam_sio_report}, highlight the need for robust dataset management in machine learning.
    
    \item \textbf{Contextual Challenges:} The context-specific nature of AI systems complicates the development of standardized weakness and use enumerations. For instance, while a system might be theoretically vulnerable to ``Prompt Injection" \cite{mitreMITREATLASx2122}, its specific application might mitigate this risk. Identical ML models used for different purposes add complexity, making model documentation crucial for understanding relevant weaknesses and uses across various contexts.
\end{itemize}

Addressing these limitations requires ongoing collaboration between AI developers, security researchers, and domain experts to create flexible, adaptive systems that accurately capture the nuances of AI weaknesses and uses across diverse applications.

\section{Research Ethics and Social Impact}

Our interdisciplinary team, consisting of AI ethicists and security researchers, navigates the intersection of AI ethics and cybersecurity, drawing inspiration from established practices like the Common Vulnerabilities and Exposures (CVE) process. We recognize the ethical imperative of mitigating harms associated with vulnerability reporting, aligning with the broader goal of enhancing model transparency and promoting the development of safer AI systems.

\subsection{Ethical Considerations}

In adherence to ethical research principles and community norms, we disclose potential challenges and ethical considerations associated with our work. No human subject experiments were conducted, and therefore an IRB was not required.

\subsection{Author Positionality}

As authors with a foundation in privacy and cybersecurity, we have witnessed the positive outcomes of the Coordinated Vulnerability Disclosure (CVD) process for software projects, both small and large. This proposal advocates for extending the CVD process to the field of machine learning research, aiming to bridge the gap and introduce machine learning researchers to the benefits of a familiar and effective process. Our positionality as proponents of CVD in this context stems from our understanding of its commercial advantages and positive impact on software security.

\subsection{Adverse Impact}

In the majority of cases, the CVD process operates smoothly without adverse consequences. However, we acknowledge instances where a vendor may reject an issue without engagement, addressing it quietly without public acknowledgment. While such occurrences are limited, they emphasize the need for a framework that encourages responsible disclosure and engagement, ensuring transparency and accountability in the machine learning community.

\bibliography{ref}

\end{document}